\documentclass[sigconf]{acmart}

\usepackage{enumitem}
\usepackage{MnSymbol} 
\usepackage{amsmath}
\usepackage{bm}
\usepackage{bbm}
\usepackage{multirow}
\usepackage{subfigure}
\usepackage{algorithm}
\usepackage{algorithmic}
\usepackage{wrapfig}

\AtBeginDocument{%
  }

\copyrightyear{2025}
\acmYear{2025}
\acmDOI{XXXXXXX.XXXXXXX}

\acmConference[ACM SIGSPATIAL 2025]{Make sure 33rd ACM SIGSPATIAL International Conference on Advances in Geographic Information Systems}{November 03--06, 2025}{Minneapolis, MN, USA}
\acmISBN{978-1-4503-XXXX-X/18/06}

\begin{document}

\title[Pre-request Waiting Time]{A First Look at Predictability and Explainability of \\Pre-request Passenger Waiting Time in Ridesharing Systems}

\author{Jie Wang}
\affiliation{
  \institution{Independent Researcher}
  % \city{Tallahassee, Florida}
  \country{United States}
}

\author{Guang Wang}
\authornote{Guang Wang is the corresponding author.}
% \email{guang@cs.fsu.edu}
\affiliation{
  \institution{Florida State University}
  % \city{Tallahassee, Florida}
  \country{United States}
}

\begin{abstract}

Passenger waiting time prediction plays a critical role in enhancing both ridesharing user experience and platform efficiency. While most existing research focuses on \textbf{post-request} waiting time prediction with knowing the matched driver information, \textbf{pre-request} waiting time prediction (i.e., before submitting a ride request and without matching a driver) is also important, as it enables passengers to plan their trips more effectively and enhance the experience of both passengers and drivers. However, it has not been fully studied by existing works. In this paper, we take the first step toward understanding the predictability and explainability of pre-request passenger waiting time in ridesharing systems. Particularly, we conduct an in-depth data-driven study to investigate the impact of demand–supply dynamics on passenger waiting time.
Based on this analysis and feature engineering, we propose FiXGBoost, a novel feature interaction-based XGBoost model designed to predict waiting time without knowing the assigned driver information. We further perform an importance analysis to quantify the contribution of each factor.
Experiments on a large-scale real-world ridesharing dataset including over 30 million trip records show that our FiXGBoost can achieve a good performance for pre-request passenger waiting time prediction with high explainability.

\end{abstract}

\begin{CCSXML}
<ccs2012>
   <concept>
       <concept_id>10002951.10003227.10003351</concept_id>
       <concept_desc>Information systems~Data mining</concept_desc>
       <concept_significance>500</concept_significance>
       </concept>
   <concept>
       <concept_id>10010147.10010257.10010293</concept_id>
       <concept_desc>Computing methodologies~Machine learning approaches</concept_desc>
       <concept_significance>300</concept_significance>
       </concept>
 </ccs2012>
\end{CCSXML}

\ccsdesc[500]{Information systems~Data mining}
\ccsdesc[300]{Computing methodologies~Machine learning approaches}

\keywords{Waiting Time, Ridesharing, Predictability, Explainability}

\maketitle

% \vspace{-3pt}
\section{Introduction}\label{introduction}

Passenger waiting time prediction is very important for ridesharing services because inaccurate waiting time estimations on mobile Apps may lead to poor user experience and potential loss of users to other platforms. 
Most existing work focuses on the \textbf{post-request waiting time} \cite{jiang2025hcride}, which can utilize information of the assigned driver (e.g., driver location and distance to passengers) for prediction. However, \textbf{pre-request waiting time} prediction is largely ignored by existing work, which is also significant for passenger and driver experience, as well as platform efficiency. 

\begin{figure}[!h]
\vspace*{-6pt}
%\hspace*{-2pt}
\includegraphics[width=0.5\textwidth, keepaspectratio=true]{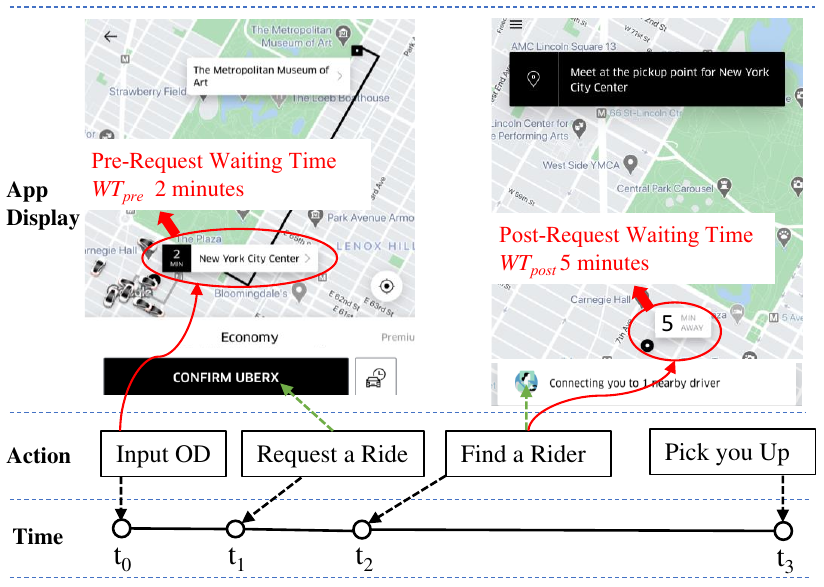}
\vspace*{-16pt}
\caption{Pre-request\&Post-request waiting time estimations.} \label{fig:AppWaitTime}
\vspace*{-6pt}
\end{figure}

Figure~\ref{fig:AppWaitTime} shows the difference between pre-request waiting time $WT_{pre}$ and post-request waiting time $WT_{post}$. 
In many situations, passengers may simply want to check whether nearby vehicles are available and how far in advance they should send a request based on the pre-request waiting time estimation.
However, our previous study \cite{wang2025mixed} reveals that significant discrepancies between pre-request and actual waiting times are common. Such inconsistencies can disrupt passengers' schedules and cause negative user experiences.
Moreover, inaccurate pre-request waiting time estimates can lead to higher costs for passengers since many ridesharing platforms impose additional charges for driver waiting time after arrival. 
For example, a per-minute waiting time fee (\$0.60/minute) will begin 2 minutes after the Uber driver arrives at the passenger pickup location \cite{UberWait}.
DiDi also charges a waiting time fee beginning 5 minutes after drivers arrive at passengers' origins for 1 CNY/minute \cite{DiDiWait}. 
Hence, this paper aims to improve the accuracy of pre-request waiting time prediction and explain the key factors influencing it using large-scale real-world ridesharing data.
The major contributions of this paper are as follows:
\begin{itemize}
% \vspace{-5pt}
    \item \textbf{Conceptually}, this is the first study of the predictability of pre-request (i.e., before submitting a ride request and without matching a driver) passenger waiting time in ridesharing systems, which is important for improving user experience and human-centered ridesharing system design.
    
    \item \textbf{Technically}, we design a novel and explainable feature interaction-based model called FiXGBoost by capturing latent feature interaction based on feature engineering and collaborative filtering for prediction.
    
    \item \textbf{Experimentally}, we evaluate our FiXGBoost using a large-scale ridesharing dataset with over 30 million records, and the results show our model can effectively outperform existing models for pre-request waiting time prediction by 28.2\%. 

\end{itemize}

\section{Data-Driven Analysis}\label{data}

Intuitively, passenger waiting time is impacted by both demand and supply. Hence, in this section, we quantitatively analyze the impact of passenger demand and vehicle supply on passenger waiting times in one of the largest ridesharing systems in the world.

\begin{figure}[!htb]
\vspace*{-10pt}
%\hspace*{-2pt}
\includegraphics[width=0.48\textwidth, keepaspectratio=true]{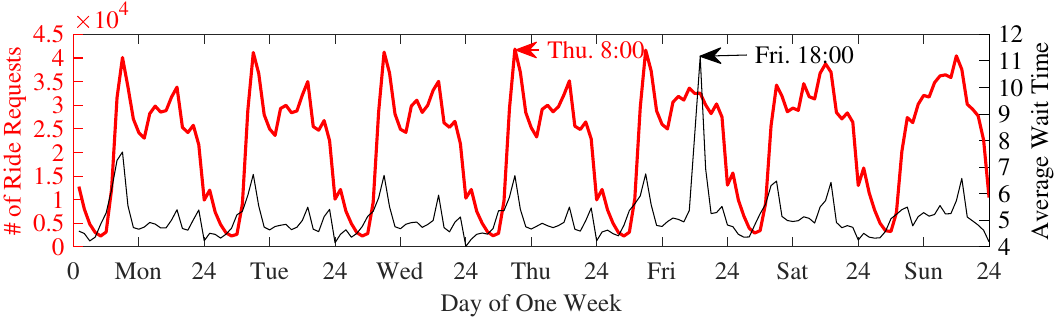}
\vspace*{-22pt}
\caption{Number of ride requests and average waiting time in each hour of a week.} \label{fig:rideWTOneWeek}
\vspace*{-10pt}
\end{figure}

Fig.~\ref{fig:rideWTOneWeek} shows the number of order requests and the average passenger waiting time in each hour of a week. We found that there is a high correlation between the number of order requests and passenger waiting time, e.g., very high passenger demand in 8:00-9:00, and the passenger waiting time is also long during this period. Surprisingly, we found there is a surge in the waiting time on Friday from 18:00-19:00. The reason would be that there is high passenger demand during the whole Friday afternoon, resulting in a shortage of ridesharing vehicles. In addition, the pattern of passenger demand at weekends is also different from weekdays, as well as the waiting time. Specifically, the passenger demand peaked at 18:00-19:00 on weekends, and the waiting time is also the highest in this period.

Considering demand alone does not fully capture waiting time patterns, as waiting times may be low under high demand if there is sufficient supply. Hence, we introduce the concepts of driver \textit{deficiency} and \textit{availability} to characterize the relationship between ridesharing demand and supply, which are defined as follows:

\vspace*{-8pt}
\begin{eqnarray}
Deficiency\left( i \right) = {{\mathcal{N}_{request}}\left( i \right)}-{{\mathcal{N}_{driver}}\left( i \right)}
\label{eq:deficiency}
\end{eqnarray}

\vspace*{-10pt}
\begin{eqnarray}
Availability\left( i \right) = \frac{{{\mathcal{N}_{driver}}\left( i \right)}}{{{\mathcal{N}_{request}}\left( i \right)}}
\label{eq:Availability}
\end{eqnarray}

% $Availability(i)$ denotes the average number of available drivers to each ride request in the $ith$ time slot of a day

\begin{figure}[!htb]
\vspace*{-5pt}
\begin{minipage}[c]{0.21\textwidth} \centering
%\vspace*{-3pt}
\hspace*{-0pt}
\includegraphics[width=1.0\textwidth, keepaspectratio=true]{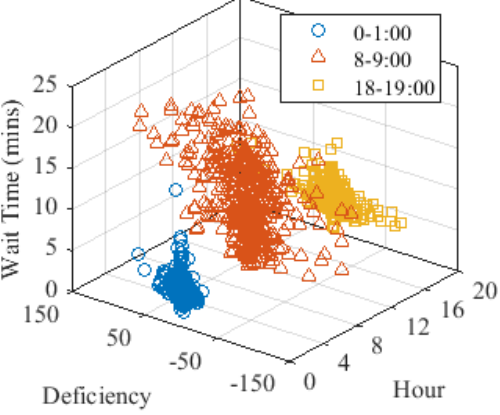}
\vspace*{-15pt}
\caption{Impact of deficiency.} \label{fig:Defi}
%\vspace*{-12pt}
\end{minipage}
\hspace*{20pt}
\begin{minipage}[c]{0.21\textwidth} \centering
%\vspace*{-3pt}
\includegraphics[width=1.0\textwidth, keepaspectratio=true]{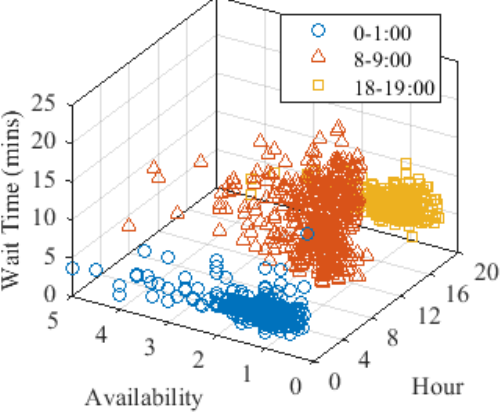}
\vspace*{-15pt}
\caption{Impact of availability.} \label{fig:Avail}
%\vspace*{-12pt}
\end{minipage}
\vspace*{-10pt}
\end{figure}

% \begin{wrapfigure}{r}{0.35\textwidth}
% \vspace*{-5pt}
% \hspace*{-5pt}
% \includegraphics[width=0.4\textwidth, keepaspectratio=true]{fig/Defi.pdf}
% \vspace*{-10pt}
% \caption{Impact of deficiency.} \label{fig:Defi}
% \vspace*{-10pt}
% \end{wrapfigure}
Where $\mathcal{N}_{driver}$ denotes the number of available drivers in the $ith$ time slot; $\mathcal{N}_{requests}$ denotes the number of ride requests in the $ith$ time slot; $Deficiency(i)$ denotes the vehicle deficiency in the $ith$ time slot of a day; $Availability(i)$ denotes the average number of available drivers to each ride request in the $ith$ time slot of a day. Due to the highly dynamic temporal characteristics of ridesharing services, we set one hour as a time slot to capture more fine-grained patterns. Furthermore, due to the highly dynamic spatial characteristics of ridesharing services, we depict the average passenger waiting time in each region in each hour. Fig.~\ref{fig:Defi} and Fig.~\ref{fig:Avail} show the correlation between deficiency \& availability and waiting time in each region in different hours of a day, e.g., late-night hour, morning rush hour, and evening rush hour. Hence, each point in Fig.~\ref{fig:Defi} and Fig.~\ref{fig:Avail} means the average waiting time in a region in an hour. 

From Fig.~\ref{fig:Defi}, we found the deficiency of 8:00-9:00 is more disperse, and there are many trips with longer waiting time during this time slot, which means there is a more unbalanced demand-supply during 8:00-9:00, resulting in longer waiting times. The deficiency in the late-night hour is more intensive, which may be caused by low passenger demand. From Fig.~\ref{fig:Avail}, we found the availability is similar during different time slots, but the waiting time is more diverse during 8:00-9:00. One possible reason could be that the randomness of waiting time in the morning rush hour is higher due to high demand and complex dispatching algorithms.

We also utilize the Fano factor \cite{charles2018dethroning} to quantify the dispersion of the waiting time distribution in different hours, which is defined as the variance of the waiting time ${\sigma _T^2}$ in each time slot $T$ divided by the mean of the waiting time ${\mu _T}$,
% denoted as $F = \frac{{\sigma _T^2}}{{{\mu _T}}}$.
as shown in Equation~\ref{eq:fano}.

\vspace*{-3pt}
\begin{eqnarray}
F = \frac{{\sigma _T^2}}{{{\mu _T}}}
\label{eq:fano}
\end{eqnarray}

In addition, we utilize the widely adopted Pearson correlation coefficient (PCC) to show the correlation of waiting time and impact factors. We leverage the p-value to show the statistical significance, e.g., a p-value less than 0.05 (typically $\leq$ 0.05) indicates that the two variables are statistically significant. 
% which is shown in Equation~\ref{eq:person}.

% \vspace*{-8pt}
% \begin{equation} \small
% {\rho _{\mathcal F, \mathcal W}} = \frac{{\sum\limits_{i = 1}^n {\left( {{\mathcal F_i} - \mathcal {\bar F}} \right)} \left( {{\mathcal W_i} - \mathcal {\bar W}} \right)}}{{\sqrt {\sum\limits_{i = 1}^n {{{\left( {{\mathcal F_i} - \mathcal {\bar F}} \right)}^2}} } \sqrt {\sum\limits_{i = 1}^n {{{\left( {{\mathcal W_i} - \mathcal{\bar W}} \right)}^2}} } }}
% \label{eq:person}
% \end{equation} 

% \noindent where $\mathcal F_i$ is the $ith$ value of impact factor $\mathcal F$; $\mathcal {\bar F}$ is the average value of impact factor $\mathcal F$; $\mathcal W_i$ is the $ith$ value of waiting time $\mathcal W$; $\mathcal{\bar W}$ is the average value of waiting time. $n$ is the total number of trips. 

From Table~\ref{tab:availability}, we found that the Fano factor of 8:00-9:00 is the largest, which means there is a large difference in waiting time in different regions, which could be caused by the high demand.

\vspace*{-7pt}
\begin{table}[h!]
\footnotesize
%\vspace*{-3pt}
\caption{Correlation of waiting time and demand-supply}
\vspace*{-5pt}
\centering
\begin{tabular}{|c|c|c|c|c|}
\hline
Demand-Supply & Time Slot & Fano Factor & PCC & p-value \\ \hline
\multirow{3}{*}{\shortstack{Correlation between \\waiting time\&deficiency}} & 0:00-1:00 & 0.329 & 0.15 & 0.0023 \\\cline{2-5}
& 8:00-9:00 & 1.7655 & 0.127 & 0.0071 \\\cline{2-5}
& 18:00-19:00 & 0.302 & 0.242 & 0 \\\hline
\multirow{3}{*}{\shortstack{Correlation between \\waiting time\&availability}} & 0:00-1:00 & 0.329 & -0.124 & 0.01 \\\cline{2-5} 
 & 8:00-9:00 & 1.765 & -0.14 & 0.0028 \\\cline{2-5} 
 & 18:00-19:00 & 0.302 & 0.006 & 0.899 \\\hline
\end{tabular}
\vspace*{-8pt}
\label{tab:availability}
\end{table}

The correlation of passenger waiting time and deficiency is positive during different time slots of a day, and the results are significant ($p < 0.01$). These results indicate that high deficiency may cause a longer waiting time. The correlation of passenger waiting time and availability is negative during different 0:00-1:00 and 8:00-9:00 with $p < 0.01$, which indicates that high availability may cause shorter waiting time in these time slots. However, the correlation between waiting time and availability during 18:00-19:00 is not significant with a large $p$.

\section{Methodology}\label{methodology}

Based on data analysis, we design an explainable feature interaction-based XGBoost~\cite{chen2016xgboost} called FiXGBoost for accurate pre-request waiting time prediction.

\vspace{-5pt}
\subsection{Feature Engineering}
\label{sec:feature}
Feature engineering is one of the most important components of machine learning, especially for explainable machine learning \cite{zheng2018feature}. Hence, to achieve high prediction accuracy and explainability, based on our previous work \cite{wang2025mixed}, we first extract a set of features that are statistically significant to the passenger waiting time. We divide all features into four categories: \textbf{spatiotemporal}, \textbf{demand-supply}, \textbf{contextual}, and \textbf{trip characteristics}. 

\textbf{Spatiotemporal} features include: (1) $isRushHour$, a fine-grained temporal feature that divides a day into four time slots, i.e., morning rush hour, evening rush hour, late night, and other times; 
(2) \textit{isWeekend}, which distinguishes between weekdays and weekends; (3) $O\_{region}$, representing the trip's origin among 491 regions; (4) $D\_{region}$, the destination region; and (5) $V\_{region}$, the location of the vehicle when dispatched. 
\textbf{Demand-supply features} consist of (6) \textit{orderNum}, the number of ride orders in a region within a time slot, and (7) \textit{vehicleNum}, the number of available vehicles in that region and time. Together, these two features reflect the concepts of demand-supply deficiency and vehicle availability. 
The \textbf{contextual feature} (8) \textit{weather} captures current weather conditions such as sunny, rainy, or typhoon. Finally, \textbf{trip characteristics} features include (9) \textit{pickDistance}, which measures the distance between the dispatched vehicle and the passenger, and (10) \textit{tripDistance}, which represents the distance from the trip’s origin to its destination.

Since the $V_{region}$ and $pickDistance$ can be known only after a vehicle is assigned to serve the passenger, it is not known when we predict ${WT}_{pre}$. Hence, we utilize the other 8 features to predict ${WT}_{pre}$ and all 10 features for ${WT}_{post}$ prediction.

\vspace{-5pt}
\subsection{FiXGBoost Model}
In addition to the 10 features, we also found some hidden interaction information that may also impact passenger waiting time, e.g., real-time traffic conditions and drivers' route preference. Inspired by this, we propose a feature interaction-based XGBoost (\textbf{FiXGBoost}) to further enhance the above-extracted 10 features for passenger waiting time prediction. FiXGBoost deeply depicts the intrinsic patterns of ridesharing operations through the different interaction methods between features.

Feature interactions are crafted as combinations of individual features. The basic idea of FiXGBoost is that it captures the latent feature interaction information based on Factorization Machines (FM) \cite{rendle2010factorization}, which are widely used for feature-based collaborative filtering tasks. We extract 5 types of interaction information to improve the original single interaction methods in FM with practical considerations. 
\textbf{(i)} We consider the interaction of spatiotemporal features through collaborative filtering. We borrow the idea from recommendation system work, e.g., we consider each $O\_{region}$ as a user and each $D\_{region}$ as an item, so relevant $O\_{region}$ has a similar number of $D\_{region}$ owing to the driving pattern. For the temporal information, we consider each $O\_{region}$ as a user and $(D\_{region}, isRushHour/isWeekend$ as an item. Similarly, we also consider $D\_{region}$ or $V\_{region}$ as a user and $O_{region}, isRushHour/isWeekend$ as an item to filter the latent information.
\textbf{(ii)} The coordinates of origins and destinations are decomposed in 3D spaces, e.g., $O_x=cos(O\_lat)*cos(O\_lng)$, $O_y=cos(O\_lat)*sin(O\_lng)$, and $O_z=sin(O\_lat)$. 
\textbf{(iii)} We represent the $O\_{region}$ and $D\_{region}$ by different interaction models including their Manhattan distance, Euclidean distance, and geographical distance. \textbf{(iv)} Collaborative filtering for driver preference. We consider each driver as an user, and the frequency of $O\_{region}$, $D\_{region}$, or $(O\_{region}, D\_{region})$ as an item, and we also consider fine-grained features $(O\_{region}/D\_{region}, isRushHour/isWeekend)$ or $(O\_{region}, D\_{region}, isRushHour/isWeekend)$ as an item for the interaction information of these features. 
\textbf{(v)} We cluster the $O\_{region}$, $D\_{region}$, $V\_{region}$ into different categories by the K-means algorithm to capture the popular regions and reduce the region count. Finally, the vehicle irrelevant feature interaction information (i.e., (i)-(iv)) is fed to models for ${WT}_{pre}$ prediction and all interaction information is used for ${WT}_{post}$ prediction.

The FiXGBoost prediction model can be represented as 

\vspace*{-10pt}
\begin{equation}
\small
\hat y^{(i)} = \sum_{k=1}^{K}h_k(\mathbf{x}^{(i)}), h_k \in \mathcal H,
\label{eq:eq2}
\end{equation}

\noindent where $K$ is the number of trees; $\mathbf{x}^{(i)}$ is the $i^{th}$ input, including the extracted features from our feature engineering and the interaction features; $\hat y^{(i)}$ is the predicted waiting time, which is learned by a tree ensemble model with a collection $\mathcal H$ of $K$ functions $h_k$. The objective function at training round $t$ iteration can be denoted as

\vspace*{-10pt}
\begin{equation}
\small
\mathcal L^{(t)} = \sum_{i}^{}(l(y^{(i)}, \hat y^{(i)})) + \sum_{k}^{} \Omega (h_k),
\label{eq:eq3}
\end{equation}

\noindent where $l(.)$ is the loss function (e.g., Square loss); $\Omega$ is the regularization term (e.g., $L2$ norm), which measures the model complexity. 

\vspace{-5pt}
\section{Evaluation}\label{evaluation}

\subsection{Dataset}
We evaluate our model with a large-scale dataset from a ridesharing operator in Shenzhen, which includes over 30 million trips in 2019. Each data record consists of fields describing the order information, e.g., the order ID, order time, assigned vehicle GPS coordinates, pick-up time, pick-up GPS coordinates, drop-off time, drop-off GPS coordinates, trip length, pick-up distance, etc.

\vspace{-5pt}
\subsection{Baselines}

We compare our FiXGBoost with several widely adopted prediction methods including Linear Regression (LR), Support Vector Regression (SVR), Random Forest (RF), Multilayer Perceptron (MLP), Graph Convolutional Network (GCN), Gradient Boosting Decision Tree (GBDT), and eXtreme Gradient Boosting (XGBoost) \cite{chen2016xgboost}.

\subsection{Performance Metrics}

We utilize two widely used metrics to quantify the prediction performance of different methods, i.e., Mean Absolute Error (MAE), and Root Mean Square Error (RMSE), defined as below.

%  Mean Square Error (MSE),
\vspace{-5pt}
\begin{eqnarray}
\small
MAE = \frac{1}{{\left| \mathcal D \right|}}\sum\limits_{i = 1}^{\left| \mathcal D \right|} {\left| {{ WT_{act}^{(i)}} - {{{WT}}^{(i)}}} \right|} 
\label{eq:MAE}
\end{eqnarray}

\vspace{-5pt}
\begin{eqnarray}
RMSE = \sqrt { \frac{1}{{\left| \mathcal D \right|}}\sum\limits_{i = 1}^{\left| \mathcal D \right|} {{{\left( {{WT_{act}^{(i)}} - {{{WT}}^{(i)}}} \right)}^2}} }
\label{eq:RMSE}
\end{eqnarray}

\noindent where $WT_{act}^{(i)}$ is the actual waiting time of $ith$ trip, and ${{WT}}^{(i)}$ is the predicted waiting time, which could be ${WT}_{pre}^{(i)}$ or ${WT}_{post}^{(i)}$. $\mathcal D$ is the test set for evaluation.

\vspace{-5pt}
\subsection{Prediction Results}

The prediction error of ${WT}_{pre}$ and ${WT}_{post}$ are shown in Table~\ref{tab:wti} and Table~\ref{tab:wtp}, respectively. We found the FiXGBoost achieves the best performance for both ${WT}_{pre}$ and ${WT}_{post}$, with only 112 and 98 of MAE, respectively. It means the average ${WT}_{post}$ prediction error for all trips is about 1.6 minutes, and the average ${WT}_{pre}$ prediction error is about 1.86 minutes. Even though with the driver information, the post-request waiting time prediction is more accurate, we can still achieve high prediction accuracy for pre-request waiting time ${WT}_{pre}$, which means it has the potential for us to achieve high accuracy for waiting time prediction without assigned driver information. It is worth noting that our FiXGBoost can make the closest prediction results for ${WT}_{pre}$ and ${WT}_{post}$, which indicates its capability for more accurate ${WT}_{pre}$ prediction.

\vspace{5pt}
\hspace*{-13pt}
% \vspace*{10pt}
\begin{minipage}{\textwidth}
\small
\begin{minipage}[t]{0.24\textwidth}
  \centering
  \captionof{table}{${WT}_{pre}$ prediction (s)} \label{tab:wti}
  \vspace*{-7pt}
  \begin{tabular}{|c|c|c|}
  \hline
  Methods & MAE & RMSE\\
  \hline
  LR & 176 & 265\\
  SVR & 216 & 285\\
  RF & 163 & 246\\
  MLP & 188 & 279\\
  GCN & 178 & 266\\
  GBDT & 165 & 250\\
  XGBoost & 156 & 238\\
  \textbf{FiXGBoost} & \textbf{112} & \textbf{136}\\
  \hline
  \end{tabular}
\end{minipage}
\begin{minipage}[t]{0.24\textwidth}
  \centering
  \captionof{table}{${WT}_{post}$ prediction (s)} \label{tab:wtp}
  \vspace*{-7pt}
  \begin{tabular}{|c|c|c|}
  \hline
  Methods & MAE & RMSE\\
  \hline
  LR & 140 & 224\\
  SVR & 272 & 214\\
  RF & 131 & 209\\
  MLP & 187 & 278\\
  GCN & 177 & 266\\
  GBDT & 132 & 211\\
  XGBoost & 126 & 200\\
  \textbf{FiXGBoost} & \textbf{98} & \textbf{132}\\
  \hline
  \end{tabular}
\end{minipage}
\end{minipage}
\vspace{3pt}

\begin{figure}[!t]
% \vspace*{-8pt}
\begin{minipage}[c]{0.21\textwidth} \centering
%\vspace*{-3pt}
\hspace*{-0pt}
\includegraphics[width=1.0\textwidth, keepaspectratio=true]{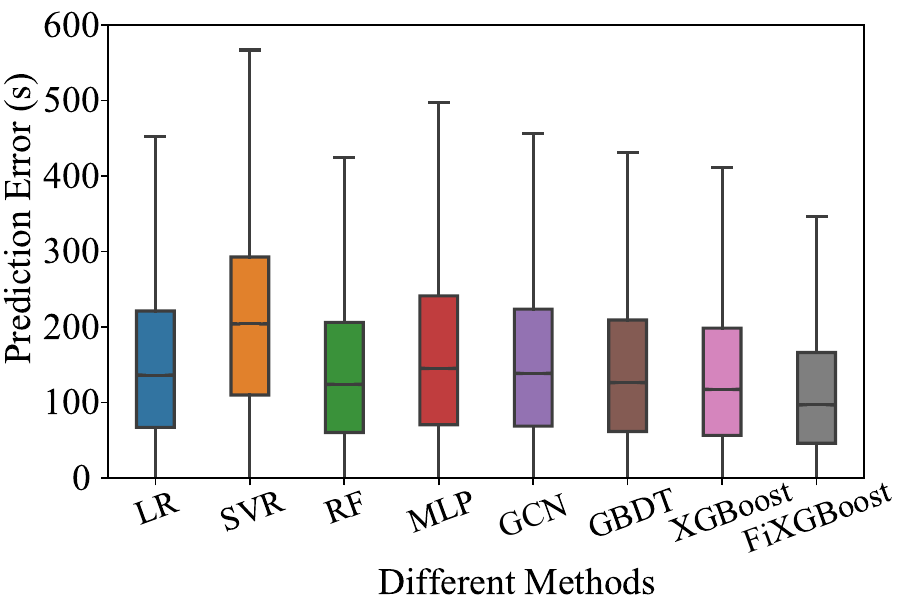}
\vspace*{-25pt}
\caption{Prediction error of ${WT}_{pre}$.} \label{fig:error2}
%\vspace*{-12pt}
\end{minipage}
\hspace*{5pt}
\begin{minipage}[c]{0.21\textwidth} \centering
%\vspace*{-3pt}
\hspace*{-0pt}
\includegraphics[width=1.0\textwidth, keepaspectratio=true]{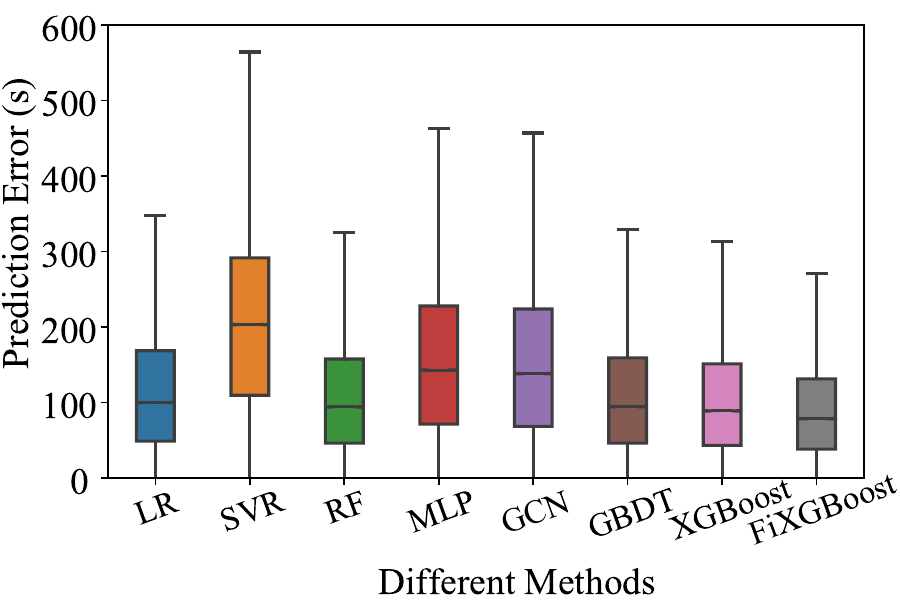}
\vspace*{-25pt}
\caption{Prediction error of ${WT}_{post}$.} \label{fig:error1}
%\vspace*{-12pt}
\end{minipage}
\vspace*{-8pt}
\end{figure}

\begin{figure}[!t]
% \vspace*{-10pt}
\begin{minipage}[c]{0.225\textwidth} \centering
%\vspace*{-3pt}
\includegraphics[width=1.0\textwidth, keepaspectratio=true]{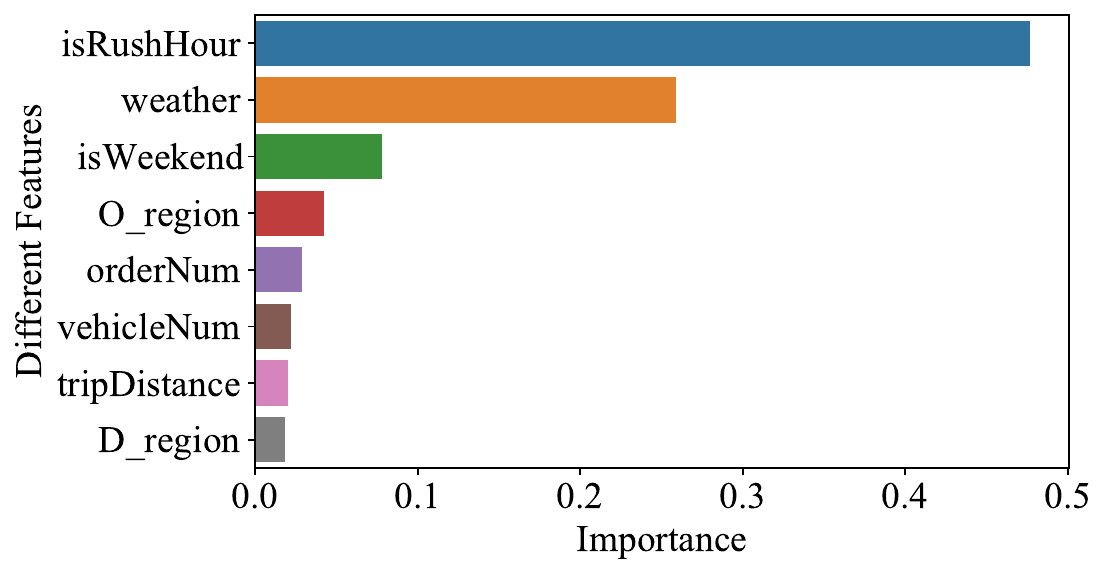}
\vspace*{-22pt}
\caption{Feature Importance for ${WT}_{pre}$ prediction.} \label{fig:xgboost2}
%\vspace*{-12pt}
\end{minipage}
\hspace*{5pt}
\begin{minipage}[c]{0.225\textwidth} \centering
%\vspace*{-3pt}
\includegraphics[width=1.0\textwidth, keepaspectratio=true]{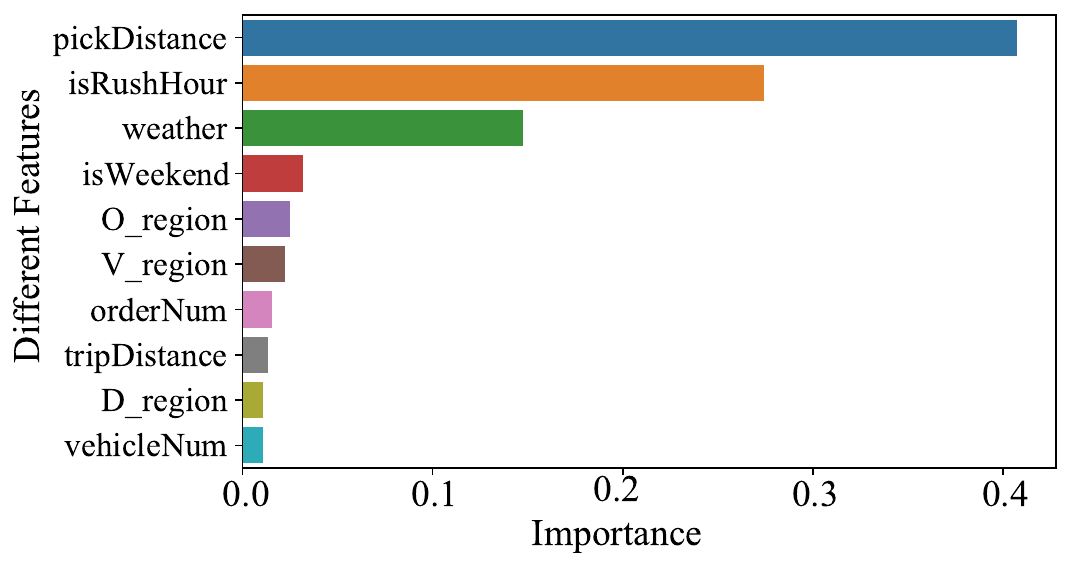}
\vspace*{-22pt}
\caption{Feature Importance for ${WT}_{post}$ prediction.} \label{fig:xgboost1}
%\vspace*{-12pt}
\end{minipage}
\vspace*{-8pt}
\end{figure}

Fig.~\ref{fig:error2} shows the pre-request waiting time ${WT}_{pre}$ prediction error of each trip, and we can see that the performance of our proposed FiXGBoost is better than other methods, and the prediction error of 65\% of trips is less than 2 minutes. Since FiXGBoost has strong interpretability, it is easy for us to obtain the importance of each feature learned by FiXGBoost. Fig.~\ref{fig:xgboost2} shows the importance of different features for ${WT}_{pre}$ prediction. We found that temporal features (e.g., \textit{isRushHour}, \textit{isWeekend}) and weather conditions are the most significant features, and the fine-grained temporal feature (i.e., \textit{isRushHour}) is more important than the time of week feature (i.e., \textit{isWeekend}). 

Fig.~\ref{fig:error1} shows the post-request waiting time prediction error of each trip, and we can see that FiXGBoost also shows the best performance, and the prediction error of 71\% of trips is less than 2 minutes. Similarly, we also obtain the importance of each feature for ${WT}_{post}$ learned by FiXGBoost, as shown in Fig.~\ref{fig:xgboost1}. We found the pickup distance (i.e., \textit{pickDistance}) is the most important feature for prediction of the waiting time ${WT}_{post}$, so it has the highest correlation with the waiting time, which is also intuitive, i.e., the vehicle travel time is highly correlated to its travel distance. The importance of other features is similar to that of ${WT}_{pre}$.

Our results show that although we can achieve accurate waiting time prediction by using driver location information, our feature interaction-based FiXGBoost can also achieve comparable performance for ${WT}_{pre}$ prediction without using driver information.

% \vspace{-5pt}
\section{Conclusion and Future Work}\label{conclusion}

In this work, we focus on a largely underexplored research topic, i.e., pre-request waiting time prediction in ridesharing systems. We first conduct an in-depth data-driven analysis to understand the correlation between passenger waiting time and demand-supply dynamics. We then extracted four categories of features that may impact pre-request waiting time prediction. 
To model these complex relationships, we propose a novel feature interaction-aware model called FiXGBoost, which leverages collaborative filtering to capture hidden feature interactions to enhance pre-request waiting time prediction accuracy.
Evaluation results on a real-world large-scale dataset from a major ridesharing platform show that our FiXGBoost can improve the prediction performance by 28.2\% compared to existing methods. We also provide an interpretable analysis of feature importance, revealing intuitive and insightful patterns.

In future work, we plan to develop more advanced prediction models based on Transformers and LLMs.

% \begin{acks}
% This material is based upon work supported by ...
% \end{acks}

% \clearpage
% \newpage
% \balance
\vspace{-5pt}
\bibliographystyle{ACM-Reference-Format}
\bibliography{reference}

\newpage
% \appendix
% \input{chap/appendix}
\end{document}